\documentclass[letterpaper, 10 pt, conference]{ieeeconf}
\IEEEoverridecommandlockouts
\let\labelindent\relax
\usepackage{enumitem}
\usepackage{balance}
\usepackage[font={small}]{caption}
\usepackage{array}
\usepackage{textcomp}
\usepackage{mathtools, nccmath}
\usepackage{graphicx}
\usepackage{amsfonts}
\usepackage{amsmath}
\usepackage{amssymb}
\usepackage{threeparttable}
\usepackage{hyperref}
\usepackage{tikz}
\usepackage{arydshln}
\usepackage{multirow}
\usepackage{bm}
\usepackage{epstopdf}
\usepackage{cite}
\usepackage{siunitx}
\usepackage{xcolor}
\usepackage[ruled,linesnumbered,boxed,noend]{algorithm2e}
\usepackage{booktabs}
\usepackage{graphicx}
\usepackage{threeparttable}
\usepackage{stfloats}
\usepackage{subfigure} 

\usepackage{subfigure}
\usepackage{graphicx}
\usepackage{balance}
\usepackage[ruled,linesnumbered,boxed,noend]{algorithm2e}
\usepackage{booktabs}
\usepackage{threeparttable}
\usepackage{hyperref}
\usepackage{array}
\usepackage{multirow}
\usepackage{textcomp}
\usepackage{stfloats}
\usepackage{url}
\usepackage{verbatim}
\usepackage{cite}
\usepackage{algpseudocode}
\usepackage{stfloats}
\usepackage{algorithmicx}
 \usepackage{amsmath}

\title{\bfseries \parbox{0.85\linewidth}{\centering COMPASS: Confined-space Manipulation Planning with Active Sensing Strategy}}
\author{ Qixuan Li$^{1,*}$, Chen Le$^{1,*}$, Dongyue Huang$^{3}$, Jincheng Yu$^{2,\dagger}$, Xinlei Chen$^{1,\dagger}$
\thanks{$^*$ \textbf{Equal Contribution.} $^\dagger$ \textbf{Corresponding Authors.}}
\thanks{$^1$ Shenzhen International Graduate School, Tsinghua University, Shenzhen, China. lqx23@mails.tsinghua.edu.cn,  le-c25@mails.tsinghua.edu.cn, chen.xinlei@sz.tsinghua.edu.cn.} 
\thanks{$^2$ Department of Electronic Engineering, and the Institute for Embodied Intelligence and Robotics, Tsinghua University, Beijing, China. yu-jc@mail.tsinghua.edu.cn. }
\thanks{$^3$ The School of Electrical and Electronic Engineering, Nanyang Technological University, Singapore. dongyue.huang@ntu.edu.sg. }
\thanks{This research was supported by National Natural Science Foundation of China (62325405), Tsinghua University Initiative Scientific Research Program, Tsinghua-Efort Joint Research Center for EAI Computation and Perception and SunRisingAI Lab, Beijing National Research Center for Information Science, Technology (BNRist), Beijing Innovation Center for Future Chips, and State Key laboratory of Space Network and Communications. This paper was supported by Guangdong Innovative and Entrepreneurial Research Team Program (2021ZT09L197) and Meituan Academy of Robotics Shenzhen.}
}

\IEEEoverridecommandlockouts

\begin{document}
\maketitle

\begin{abstract}

Manipulation in confined and cluttered environments remains a significant challenge due to partial observability and complex configuration spaces. Effective manipulation in such environments requires an intelligent exploration strategy to safely understand the scene and search the target. In this paper, we propose COMPASS, a multi-stage exploration and manipulation framework featuring a manipulation-aware sampling-based planner. First, we reduce collision risks with a near-field awareness scan to build a local collision map. Additionally, we employ a multi-objective utility function to find viewpoints that are both informative and conducive to subsequent manipulation. Moreover, we perform a constrained manipulation optimization strategy to generate manipulation poses that respect obstacle constraints. To systematically evaluate method's performance under these difficulties, we propose a benchmark of confined-space exploration and manipulation containing four level challenging scenarios. Compared to exploration methods designed for other robots and only considering information gain, our framework increases manipulation success rate by 24.25\% in simulations. Real-world experiments demonstrate our method's capability for active sensing and manipulation in confined environments.
\end{abstract}

\section{Introduction}
\label{sec:Introduction}

Manipulation in confined and cluttered environments remains a significant challenge. In such spaces, manipulator's effectiveness is fundamentally challenged due to perception occlusion and kinematic constraints. On one hand, severe \textbf{perception occlusion} to target from surrounding obstacles makes methods assuming full observability\cite{chi2023diffusion} \cite{pan2025omnimanip} not work, necessitating active exploration to incrementally build an understanding of the scene. On the other hand, obstacles compounded by the manipulator's own embodiment, impose tight \textbf{kinematic constraints} that drastically limit the robot's reachable workspace, requiring that any planned motion and manipulation must be collision-free with respect to environmental obstacles. This dilemma demands a paradigm shift from simple perception-then-motion pipelines in manipulation tasks to a truly integrated perception-motion-manipulation approach.

Current paradigms for robotic manipulation fall short of addressing this challenge. End-to-end learning methods, for instance, often rely on a fixed, third-person camera for a global understanding of the scene~\cite{kim2024openvla, ze20243d}, a setup that is untenable in confined spaces where such a view is unavailable. They also learn a policy from demonstration data, which typically lacks complex obstacle interactions, making the resulting policies struggle to handle tasks in the presence of obstacles. Additionally, existing planning-based methods typically assume a complete world model and a known target pose~\cite{huang2024rekep, pan2025omnimanip}, ignoring partial observation problem. They focus on finding a path to a pre-defined goal in a totally known environment, rather than on the active information-gathering process required to first explore the space and discover the goal. Both approaches fail to address the critical challenge of how a manipulator incrementally explores and understands a confined space to enable manipulation. To the best of our knowledge, there is no existing work that presents an integrated perception-motion-manipulation approach for confined-space manipulation tasks.

\begin{figure}[t]
    \centering
    \includegraphics[width=8.6cm]{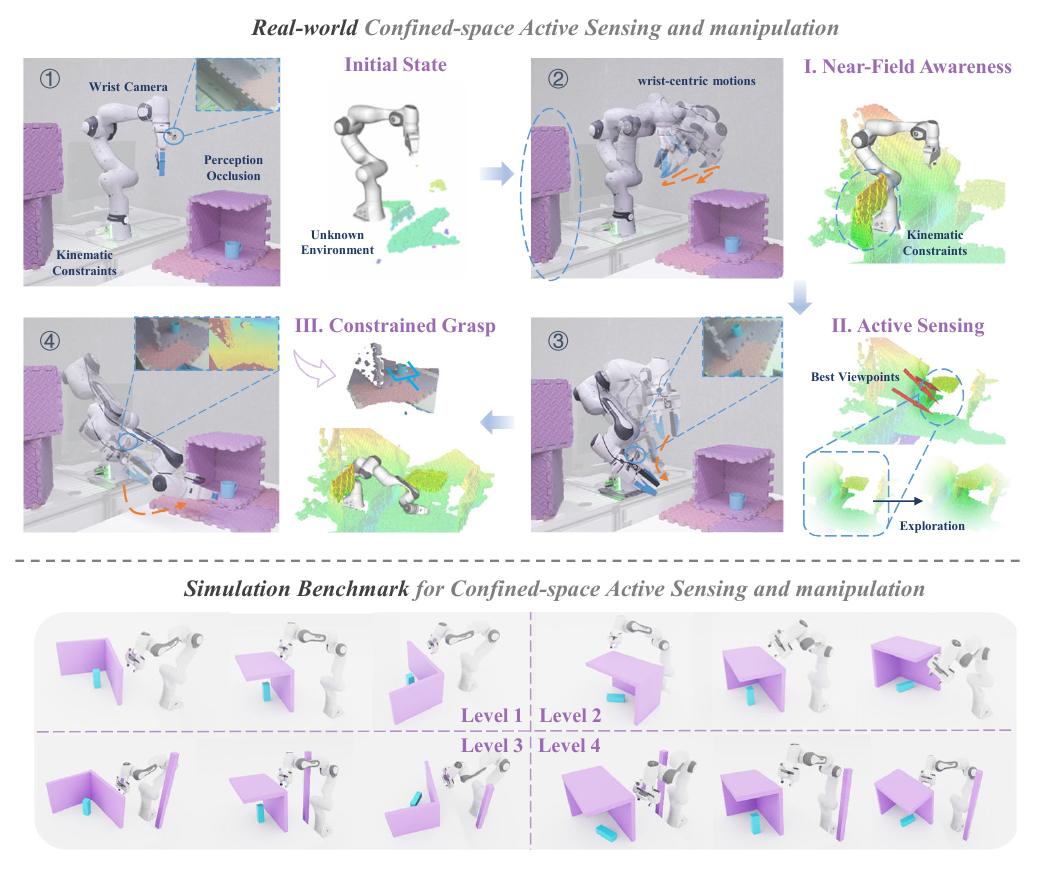}
    \caption{\textbf{Overview of COMPASS}, a framework for active perception and manipulation in confined spaces. (Top) The three-stage pipeline solves a real-world task by sequentially performing: (I) a Near-Field Awareness Scan for safety, (II) MUE-RRT to find the target, and (III) Constrained Grasp optimization.  (Bottom) Our principled simulation benchmark systematically evaluates performance across four levels of increasing difficulty of perception occlusion and kinematics constraints.}
    \label{fig:overview}
    \vspace{-0.5cm}
\end{figure}

To tackle these challenges, we propose COMPASS, a multi-stage framework for exploration and constrained manipulation. It comprises three key stages: a Near-Field Awareness Scan, a Manipulation-Utility Exploration RRT (MUE-RRT) planner, and a constrained manipulation pose generation module. The framework begins with the near-field awareness scan, which executes a series of cautious, wrist-centric motions to build a local collision map. This initial step mitigates the risks of moving in an unknown environment and increases the kinematic feasibility of motion. Once this local map is established, the system employs MUE-RRT to iteratively select the next-best-viewpoint to eliminate perception occlusion. The selection is guided by a multi-objective utility function that balances information gain, manipulability, motion cost, and task heuristics. Concurrently, an asynchronous detection process operates on an observation buffer to identify the target throughout the exploration. Once the target is identified, we perform  kinematic-constraint aware manipulation optimization  strategy to generates grasp poses that respect all obstacle constraints, leveraging the environmental understanding acquired during the exploration stage.

The proposed framework is validated on a principled and progressively challenging benchmark designed to test the task of manipulation in confined space. Simulations
and real-world experiments validate our method’s capability to perform confined-space manipulation in typical complex scenarios.  Our contributions can be
summarized as:
\begin{enumerate}
    \item We propose COMPASS, a framework for exploration and manipulation in confined spaces. It enables a manipulator with partial observation to safely explore unknown environment and perform constrained manipulation.

    \item An exploration planner for manipulators, manipulation-utility exploration RRT (MUE-RRT), is proposed to perform manipulation-centric exploration and scene understanding. It's capable of generating a series of smooth and safe exploration trajectory for manipulation tasks.
    
    \item We introduce a principled and progressively challenging benchmark for confined space manipulation tasks. We demonstrate the effectiveness of the proposed method in simulation and real-world experiments.
    
\end{enumerate}

\section{Related Works}
\label{sec:related_works}

\subsection{Embodied Manipulation}
Prevailing approaches to embodied manipulation can be categorized into end-to-end learning-based and planning-based methods. Imitation learning methods, such as VLA\cite{kim2024openvla} and Diffusion Policy\cite{chi2023diffusion}\cite{ze20243d}, directly learn manipulation policies from human demonstrations. Recent foundation model-based frameworks have demonstrated impressive embodied spatial reasoning and slow-thinking capabilities \cite{zhao2025embodied}. However, these methods are often constrained by demonstration data, which typically lacks complex obstacle interactions. And their reliance on a static, third-person camera perspective makes them ill-suited for confined spaces. Similarly, planning-based manipulation methods \cite{huang2024rekep}\cite{pan2025omnimanip} tend to focus on generating the end-effector movement and assume a complete world model and a known target
pose. They often overlook the kinematic and environmental constraints imposed on the manipulator's trajectory to reach that pose.

\subsection{Space Exploration}
Exploration has been extensively studied in the robotics community. A body of research has focused on Unmanned Aerial Vehicles (UAVs) and ground vehicles 
\cite{dharmadhikari2020motion, shah2021rapid}. 
In these studies, the methods often follow a Next-Best-View (NBV) paradigm, aiming to reduce environmental uncertainty by maximizing information gain \cite{zhu2021dsvp, cao2021tare}. 
Typically, autonomous exploration involves three key stages:
generation of candidates viewpoints/trajectories, utility evaluation, motion planning and execution\cite{placed2023survey}. Directly applying these methods to manipulator-based exploration is challenging, as it is unsafe for a manipulator to perform large-scale movements without sufficient prior understanding of its surrounding environment. Existing works for manipulator exploration\cite{naazare2022online}\cite{ren2023robot}
often focus on geometric coverage or 3D reconstruction but overlook the necessity of finding viewpoints that are conducive to subsequent grasp execution.

\subsection{Grasping in Confined Spaces}
Generating a feasible grasp in a confined space is a multi-faceted challenge. While significant progress has been made in grasp pose generation~\cite{fang2020graspnet, fang2023anygrasp}, these methods often decouple grasp quality from the robot's kinematic feasibility, leading to unreachable grasps in cluttered scenes. To address this, the field has moved towards joint grasp and motion planning, with works
~\cite{kang2022grasp}~\cite{elliott2016making} find a grasp and a path simultaneously in confined space. However, this line of work typically assumes a complete world model, sidestepping the challenge of exploration under uncertainty.

\begin{figure*}[ht]
    \centering
    \includegraphics[width=1\linewidth]{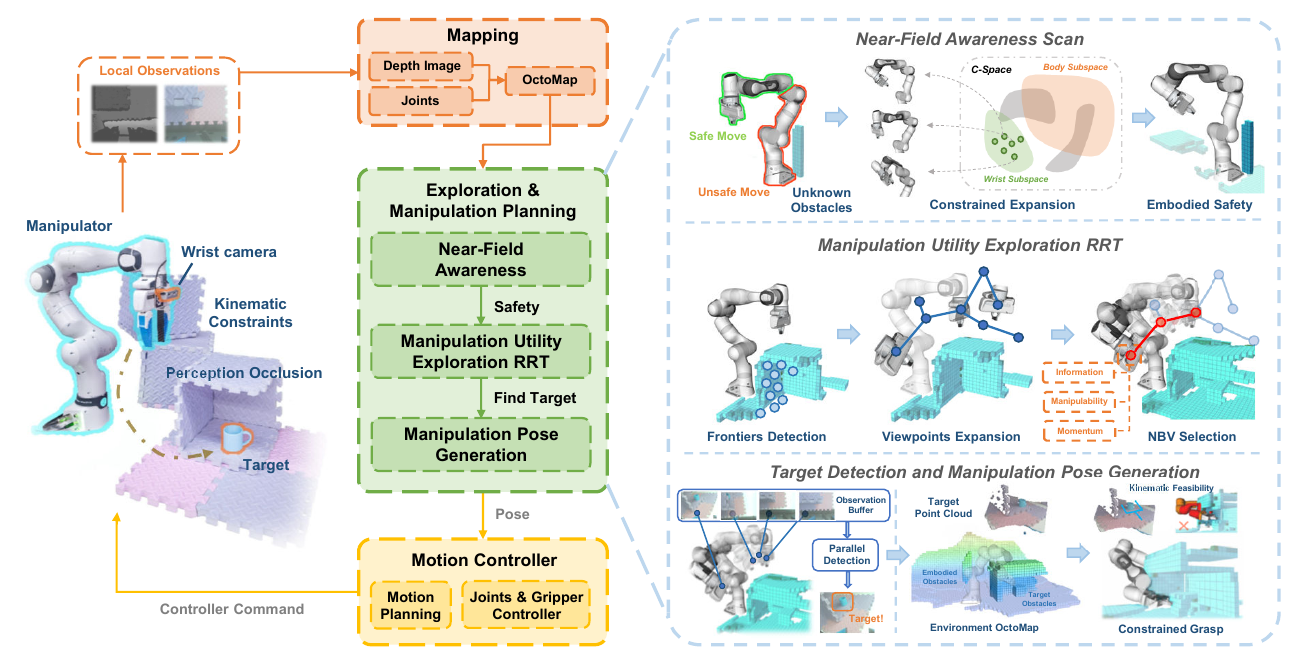}
    \caption{ Overview of system framework of COMPASS. Depth images from the wrist-camera are used to incrementally build an OctoMap. The core multi-stage planner first ensures safety with a Near-Field Awareness Scan, then executes the Manipulation-Utility Exploration RRT to conduct active perception and find the target. Upon target discovery, the Manipulation Pose Generation stage computes a whole-body, collision-free grasp. The Motion Controller plans and executes the trajectory.
    }
    \label{fig:framework}
    \vspace{-0.5cm}
\end{figure*}

\section{Overview Of The Framework}
\label{sec:framework_overview}

\subsection{Problem Formulation}
\label{sec:problem_formulation}

Define $\mathcal{W} \subset \mathbb{R}^3$ as the work space to be explored and manipulated. Let $\mathcal{W}_{free} \subset \mathcal{W}$ be the no-obstacle subspace. Define viewpoint $\boldsymbol{\mathrm{v}}\in \mathrm{SE(3)}$ to describe the pose of the camera onboard the robot,
$\boldsymbol{\mathrm{v}} = [\boldsymbol{\mathrm{t}}_{\boldsymbol{\mathrm{v}}}; \boldsymbol{\mathrm{R}}_{\boldsymbol{\mathrm{v}}}]$ where $\boldsymbol{\mathrm{t}}_{\boldsymbol{\mathrm{v}}}\in\mathcal{W}_{free}$ and $\boldsymbol{\mathrm{R}}_{\boldsymbol{\mathrm{v}}} \in \mathrm{SO(3)}$ respectively
denote the position and orientation. Define $\mathcal{Q} \subset \mathbb{R}^{d}$ as the configuration space (C-space), where $d$ is the DOF of manipulator. Define manipulation pose $\boldsymbol{\mathrm{e}}\in \mathrm{SE(3)}$ to describe the pose of the end effector. Our problem can be formulated as follows. 

\textit{Exploration problem for manipulator in each cycle:} Given current joint configuration $\boldsymbol{q}_{current}\in\mathcal{Q}$ and map $\mathcal{M}_{current}$, find optimal exploration path $\mathcal{T}^*=[\boldsymbol{\mathrm{v}}_1,\boldsymbol{\mathrm{v}}_2,...]$, which corresponding to a feasible trajectory in C-space $\boldsymbol{\tau}=\{\boldsymbol{q}(t)|t=[0,T]\}$. $\mathcal{T}^*$ is optimal means that it is shortest and cover more surface, $\boldsymbol{\tau}$ is feasible means that it is collision-free and avoids singular configurations.

The above problem is solved iteratively to select viewpoints and plan trajectories. When target is found, a manipulation pose $\boldsymbol{\mathrm{g}}=(\boldsymbol{\mathrm{e}},w)$ is generated, where $\boldsymbol{\mathrm{e}}$ is the pose of gripper and $w\in\mathbb{R}$ denotes the gripper width.


\subsection{System Framework}
\label{sec:system_framework}

In this paper, we propose a framework for active perception and manipulation in confined spaces, illustrated in Fig.~\ref{fig:framework}. Firstly, a map node incrementally builds an octomap from wrist-camera depth images. Then, an exploration node works in a three-stage manner: (1) Near-Field Awareness Scan serves for embodiment safety; (2) Manipulation-Utility Exploration  constructs the environment understanding and detects the target; (3) Manipulation pose generation stage
generates the grasp pose in confined space. A motion node communicates with the manipulator and controls it. 

\section{Methods}
\label{sec:methods}

\subsection{Manipulation-Oriented Exploration}

Manipulation in confined spaces presents the dual challenges of perception occlusion and kinematic constraints. Obstacles near the manipulator pose significant collision risks, while those surrounding the target cause severe perception occlusions, transforming the task into an active perception and exploration problem. To address this, we propose a multi-stage exploration strategy, the Manipulation-Utility Exploration RRT (MUE-RRT), which operates in two stages: a near-field awareness stage to ensure immediate safety, and a global stage to efficiently detect the target.

\subsubsection{Near-Field Awareness Scan}

To mitigate collision risks in the initially unknown environment, our framework begins with a Near-Field Awareness Scan. The goal is to find a minimal set of joint configurations that maximizes the volumetric coverage of a predefined safety envelope.

We formulate this as a constrained viewpoint set optimization problem. First, we define a target safety volume, $\Omega_{\text{safe}}$, as a bounding box around the robot's base. The robot's body configuration, $\boldsymbol{q}_b \in \mathcal{Q}_B$ (i.e., the base and shoulder joints), is held constant at its safe initial posture, $\boldsymbol{q}_{b,init}$. The optimization is then performed over the wrist's configuration space, $\mathcal{Q}_W$ (i.e., the wrist joints). We generate a set of candidate wrist configurations, $\mathcal{Q}_{W}^{\text{cand}}$, by uniformly sampling this subspace.

The final waypoints, $\{\boldsymbol{q}_{w,1}^*, \dots, \boldsymbol{q}_{w,k}^*\} \subset \mathcal{Q}_W$, are computed greedily and then executed sequentially. At each selection step $i$, we find the wrist configuration that covers the largest remaining unknown volume within $\Omega_{\text{safe}}$:
\begin{equation} \label{eq:greedy_scan_constrained}
    \boldsymbol{q}_{w,i+1}^* = \mathop{\arg\max}\limits_{\boldsymbol{q}_w \in \mathcal{Q}_{W}^{\text{cand}}}\  \text{Volume}(\Omega_{\text{visible}}(\boldsymbol{q}_{b,init}, \boldsymbol{q}_w) \cap \Omega_{\text{safe}}^{\text{unk}, i})
\end{equation}
where the visible volume $\Omega_{\text{visible}}(\boldsymbol{q}_{b,init}, \boldsymbol{q}_w)$ is explicitly a function of the fixed body configuration $\boldsymbol{q}_{b,init}$ and the variable wrist configuration $\boldsymbol{q}_w$. $\Omega_{\text{safe}}^{\text{unk}, i}$ represents the yet-unseen portion of the safety volume at step $i$.

\subsubsection{Global Exploration}
The system's perception input is a depth image from the wrist-mounted camera. This image is registered to the world frame using forward kinematics and integrated into the global OctoMap~\cite{hornung2013octomap}, denoted as $\mathcal{M}=\{ \mathcal{V}_{free}, \mathcal{V}_{occ}, \mathcal{V}_{unk}\}$. To prevent the robot's own geometry from generating spurious obstacles, its links are dynamically filtered from $\mathcal{V}_{occ}$.

We identify frontiers in the OctoMap, which serve as the primary guidance for exploration. A frontier is defined as the boundary between known free space and unknown space~\cite{zhu2021dsvp}. Formally, a voxel is considered a frontier if it resides in free space and is adjacent to at least one unknown voxel. Detected frontier voxels are then clustered into distinct regions $\{\mathcal{F}_1, \mathcal{F}_2, \dots, \mathcal{F}_N\}$.

At each decision-making step, we dynamically build a Rapidly-exploring Random Tree (RRT)~\cite{orthey2023sampling}. Each node in the RRT corresponds to a reachable camera viewpoint, and each connecting edge represents a collision-free trajectory in the workspace. This approach significantly reduces the time required for motion planning during tree expansion. To guide the exploration efficiently, the RRT's growth is biased towards the frontier regions $\{\mathcal{F}_i\}$ and any task-relevant priors. Specifically, our sampling strategy directs new samples towards either the frontiers or the task priors with a certain probability, while performing uniform random sampling otherwise.

Once the RRT is constructed, the Next-Best-View (NBV) is selected by evaluating each node $\boldsymbol{x}=(\boldsymbol{\mathrm{v}},\boldsymbol{q})$ in the tree using a multi-objective utility function, $\mathcal{U}(\boldsymbol{x})$. This function is designed to holistically assess a candidate viewpoint by normalizing its potential rewards against its associated motion cost, thereby promoting an efficient exploration strategy that maximizes the value gained per unit of effort. The optimal next node, $\boldsymbol{x}^*$, is the one that maximizes this utility:
\begin{equation} \label{eq:utility_full}
\boldsymbol{x}^{*} =\mathop{\arg\max}\limits_{\boldsymbol{x} \in \text{RRT}}\ \mathcal{U}(\boldsymbol{x})
\end{equation}

The utility function $\mathcal{U}(\boldsymbol{x})$ is formulated as a sum of cost-normalized reward terms:
\begin{equation} \label{eq:utility_components}
\mathcal{U}(\boldsymbol{x}) = \boldsymbol{w}^T[\mathcal{G}(\boldsymbol{x}) \ \mathcal{D}(\boldsymbol{x})\  \mathcal{M}(\boldsymbol{x}) \ \mathcal{H}(\boldsymbol{x})]^T/\mathcal{C}(\boldsymbol{x})
\end{equation}
where $\boldsymbol{w}=[w_g\ w_d \ w_m \ w_h]^T$ represents the respective weighting factors for each reward component. These components are defined as follows:

\textit{1) Information Gain} $\mathcal{G}(\boldsymbol{x})$: The information gain measures the volume of new space observable from viewpoint $\boldsymbol{\mathrm{v}}$. It is defined as the number of previously unknown voxels within the viewpoint's field of view (FOV). Formally, $\mathcal{G}(\boldsymbol{x}) = \int_{v \in \mathcal{V}_{\text{vis}} \cap \mathcal{V}_{\text{unk}}} dv$, where $\mathcal{V}_{\text{vis}}$ is the set of voxels visible from $\boldsymbol{\mathrm{v}}$. This value is computed by raycasting from the virtual camera pose into the current OctoMap~\cite{dang2020graph}.

\textit{2) Exploration Momentum} $\mathcal{D}(\boldsymbol{x})$: To encourage continuous and smooth exploration paths, this term rewards viewpoints that maintain the current direction of exploration. It is calculated as $\mathcal{D}(\boldsymbol{x}) = (\mathbf{t}_{\text{prev}}-\mathbf{t}_{\text{root}}) \cdot (\mathbf{t}_{\boldsymbol{x}}-\mathbf{t}_{\text{root}})$, where $\mathbf{t}_{\boldsymbol{x}}$ is the translational component of the viewpoint, $\mathbf{t}_{\text{root}}$ is that of the RRT's root node, and $\mathbf{t}_{\text{prev}}$ corresponds to the previously selected viewpoint.
    
\textit{3) Manipulability} $\mathcal{M}(\boldsymbol{x})$: This term promotes configurations that are far from singularities, ensuring the robot remains poised for future manipulation tasks. It is quantified by the Yoshikawa manipulability index~\cite{yoshikawa1985manipulability}, $\mathcal{M}(\boldsymbol{q}) = \sqrt{\det(\boldsymbol{J}(\boldsymbol{q})\boldsymbol{J}(\boldsymbol{q})^T)}$, where $\boldsymbol{J}(\boldsymbol{q})$ is the manipulator's Jacobian matrix.

\textit{4) Motion Cost} $\mathcal{C}(\boldsymbol{x})$: This represents the total effort required to reach configuration $\boldsymbol{q}$ from the current root state $\boldsymbol{q}_{\text{current}}$. It is calculated as the path length in the robot's joint space, $\mathcal{C}(\boldsymbol{x}) = \int_{0}^{1} ||\dot{\boldsymbol{\tau}}(s)||\,ds$, where $\boldsymbol{\tau}(s)$ is the joint-space trajectory from $\boldsymbol{q}_{\text{current}}$ to $\boldsymbol{q}$. Joint-space distance provides a more accurate measure of robot effort than its Cartesian counterpart.

\textit{5) Heuristic Guidance}: We integrate task-driven guidance into the core exploration mechanism. This guidance is represented as a set of heuristic 3D points of interest, $\mathcal{P}_h$. This information is incorporated into our planner at two synergistic levels. First, at the sampling stage, we bias the RRT tree expansion process. Second, at the evaluation stage, we enhance the utility function with a heuristic gain term, $\mathcal{H}(\boldsymbol{x})$, defined as: $\mathcal{H}(\boldsymbol{x}) = \max_{\boldsymbol{p}_j \in \mathcal{P}_h} \left( \boldsymbol{\phi}_{\mathrm{v}} \cdot (\boldsymbol{p}_j - \mathbf{t}_{\mathrm{v}})/||\boldsymbol{p}_j - \mathbf{t}_{\mathrm{v}}|| \right) $, where $\boldsymbol{\phi}_{\mathrm{v}}$ is the viewing axis of the viewpoint $\mathrm{v}$, $\boldsymbol{p}_j$ is a heuristic point of interest, and $\mathbf{t}_{\mathrm{v}}$ is the translational position of the viewpoint.

\begin{algorithm}[t]
\SetAlgoLined
\DontPrintSemicolon
\KwIn{Initial robot configuration $\boldsymbol{q}_{init}$}
\KwOut{Map $\mathcal{M}_T$ and Manipulation Result $\boldsymbol{\Gamma}$}

$\boldsymbol{q}_{current} \gets \boldsymbol{q}_{init}$\;
$\{\boldsymbol{q}^*_{w}\}_{i=1:m} \gets \text{CalculateWristWaypoints}(\boldsymbol{q}_{init})$\;
$\mathcal{M}_{0} \gets \text{ConductNearFieldAwareness}(\{\boldsymbol{q}^*_{w}\}_{i=1:m})$\;
$found \gets \textbf{false}$\;

\While{$\neg\ found$}{
    $I_t \gets \text{SenseFromWristCamera}(\boldsymbol{q}_{current})$\;
    $\mathcal{M}_{t} \gets \text{UpdateMap}(\mathcal{M}_{t-1}, I_t, \boldsymbol{q}_{current})$\;
    $\mathcal{M}_{t} \gets \text{FilterRobotBody}(\mathcal{M}_{t}, \boldsymbol{q}_{current})$\;
    $\mathcal{F}_{list} \gets \text{DetectAndClusterFrontiers}(\mathcal{M}_t)$\;
    $(\mathcal{V},\mathcal{E}) \gets \text{ExpansionRRT}(\mathcal{M}_t, \mathcal{F}_{list})$\;
    $\mathcal{U} \gets \text{ComputeUtility}((\mathcal{V},\mathcal{E}),\mathcal{M}_t)$\;
    $\boldsymbol{q}_{next} \gets \text{SelectNBV}(\mathcal{U})$\;
    
    $\boldsymbol{\tau} \gets \text{PlanningPath}(\boldsymbol{q}_{current}, \boldsymbol{q}_{next})$\;
    $\boldsymbol{q}_{current} \gets \text{ExecuteTrajectory}(\boldsymbol{\tau})$\;
    $\mathcal{B}_t\gets \{ (I_{t-k}, \boldsymbol{\mathrm{v}}_{t-k}), \dots, (I_{t-1}, \boldsymbol{\mathrm{v}}_{t-1}), (I_t, \boldsymbol{\mathrm{v}}_t) \}$\;
    $found,\boldsymbol{\mathrm{v}}^* \gets \text{TargetDetection}(\mathcal{B}_t)$\;
}

$\mathcal{G} \gets \text{ManipulationPoseGeneration}(I_t,\boldsymbol{\mathrm{v}}^*,\mathcal{M}_t)$\;
$\boldsymbol{\Gamma} \gets \text{ConductManipulation}(\mathcal{G},\mathcal{M}_t)$\;
\Return $\mathcal{M}_T, \boldsymbol{\Gamma}$
\caption{Exploration and Manipulation}
\label{alg:exploration}
\end{algorithm}

\subsection{Target Detection and Manipulation Pose Generation}

To facilitate target detection during manipulator motion, our framework employs an asynchronous detection process. This module addresses the latency mismatch between the robot's continuous motion and the computationally intensive inference of detection models, such as the YOLO~\cite{cheng2024yolow} model used in this paper. The detection process maintains a spatio-temporal observation buffer, $\mathcal{B}_t$, which stores a history of synchronized image-pose pairs over a sliding time window:
\begin{equation} \label{eq:obs_buffer}
    \mathcal{B}_t = \{ (I_{t-k}, \boldsymbol{\mathrm{v}}_{t-k}), \dots, (I_{t-1}, \boldsymbol{\mathrm{v}}_{t-1}), (I_t, \boldsymbol{\mathrm{v}}_t) \}
\end{equation}
where $I_i$ is the RGB image from the wrist camera and $\boldsymbol{\mathrm{v}}_i$ is the corresponding camera pose at time $i$. The parameter $k$ denotes the buffer size. By processing this buffer in batches, the detection process can perform robust, temporally-consistent inference, ensuring that transient views of the target are not missed due to processing delays. To ensure high-fidelity grasp perception, we select the optimal observation $(I^*, \boldsymbol{\mathrm{v}}^*)$ from the subset of the buffer containing successful detections, $\mathcal{B}_{\text{det}} \subseteq \mathcal{B}_t$. This selection is based on a perceptual quality score, $S_{\text{obs}}$, defined as:
\begin{equation} \label{eq:quality_score}
    S_{\text{obs}}(I_i) = -\| \boldsymbol{p}_{\text{bbox}}(I_i) - \boldsymbol{p}_{\text{center}} \|_2
\end{equation}
where $\boldsymbol{p}_{\text{bbox}}(I_i)$ is the center of the target's 2D bounding box in image $I_i$, and $\boldsymbol{p}_{\text{center}}$ is the image center. This score prioritizes the viewpoint where the target is most centrally located, thereby maximizing view quality for the subsequent grasp pose generation.

The grasp pose is generated at the optimal viewpoint $\boldsymbol{\mathrm{v}}^*$. In this paper, we employ the GraspNet~\cite{fang2020graspnet} detection network, which processes the depth data from image $I^*$ to generate a set of candidate grasps, $\mathcal{G}_{\text{cand}} = \{ \boldsymbol{\mathrm{g}}_1, \boldsymbol{\mathrm{g}}_2, \dots, \boldsymbol{\mathrm{g}}_m \}$. Each candidate grasp $\boldsymbol{\mathrm{g}}_i$ is defined as a tuple $\boldsymbol{\mathrm{g}}_i = (\boldsymbol{\mathrm{e}}_{i}, w_i, s_i)$, where $\boldsymbol{\mathrm{e}}_{i} \in \mathrm{SE(3)}$ is the end-effector pose, $w_i$ is the gripper width, and $s_i \in (0, 1]$ is the grasp quality score. This score $s_i$ reflects only the geometric quality of the grasp (e.g., force closure as in~\cite{fang2020graspnet}) and remains naive to a critical aspect: the kinematic feasibility of the manipulator.

To address this limitation, we perform a pose refinement for each grasp pose candidate. This process can be viewed as a constrained optimization problem:
\begin{equation} \label{eq:grasp_optimization}
\begin{aligned}
\max_{\boldsymbol{\mathrm{g}}_i \in \mathcal{G}_{\text{cand}}} \quad & s_i  \\
\textrm{s.t.} \quad & \text{IK}(\boldsymbol{\mathrm{e}}_{i}) \in \mathcal{Q}_{free} \\
& \angle(\boldsymbol{z}_{\boldsymbol{\mathrm{g}}_i}, \boldsymbol{n}) \le \delta \\
& \text{FK}(\boldsymbol{\tau}(\boldsymbol{q}_{cur},\boldsymbol{q}_{grasp}))\subset \mathcal{M}_{free}
\end{aligned}
\end{equation} 
where $s_i$ defines the analytic force-closure quality evaluated under varying friction coefficients\cite{fang2020graspnet}. The term $\text{IK}(\boldsymbol{\mathrm{e}}_{i})$ denotes the inverse kinematics solution for the grasp pose. The second constraint enforces a context-aware approach direction for the gripper, where $\boldsymbol{z}_{\boldsymbol{g}_i}$ represents the approach vector corresponding to the i-th grasp candidate $\boldsymbol{g}_i$ (typically the z-axis of end effector frame), and the desired approach vector $\boldsymbol{n}$ is determined by analyzing the object's orientation. Finally, the entire workspace trajectory, obtained by applying forward kinematics $\text{FK}(\cdot)$ to the planned joint-space path $\boldsymbol{\tau}(\boldsymbol{q}_{cur},\boldsymbol{q}_{grasp})$, is required to be within the free space of the current map, $\mathcal{M}_{free}$.


\subsection{Runtime Logic}

\begin{figure}[t]
    \centering
    \includegraphics[width=8.6cm]{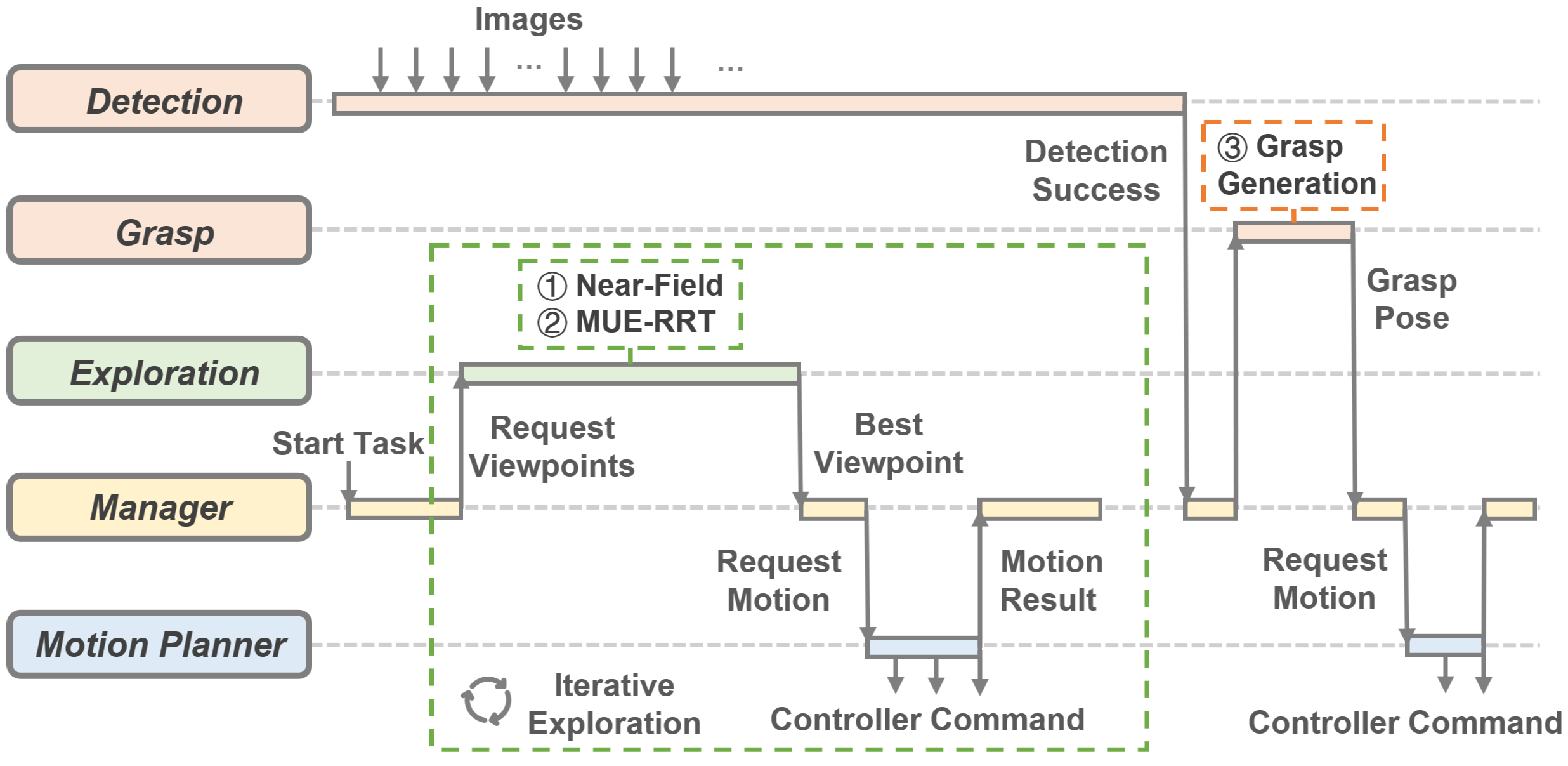}
    \caption{The runtime logic of a successful exploration and manipulation in confined space. }
    \label{fig:logic}
    \vspace{-0.5cm}
\end{figure}

The overall system is implemented as a set of interacting, asynchronous nodes for exploration, motion planning, target detection, and grasp generation, all coordinated by a manager node using a Finite State Machine (FSM). The runtime logic is illustrated in Fig.~\ref{fig:logic}. Upon receiving a \textit{start task} signal, the manager initiates an iterative exploration loop by requesting the next-best-viewpoint from the exploration node. It then commands the motion planner to generate and execute a collision-free trajectory to that viewpoint. This perception-action loop continues, with the detection node asynchronously processing incoming images, until the target is found. Upon successful detection, the manager transitions to the final phase: it requests a refined grasp pose from the grasp generation node and commands the motion planner to execute the grasp, completing the task.

\begin{figure}[t]
    \centering
    \includegraphics[width=8.6cm]{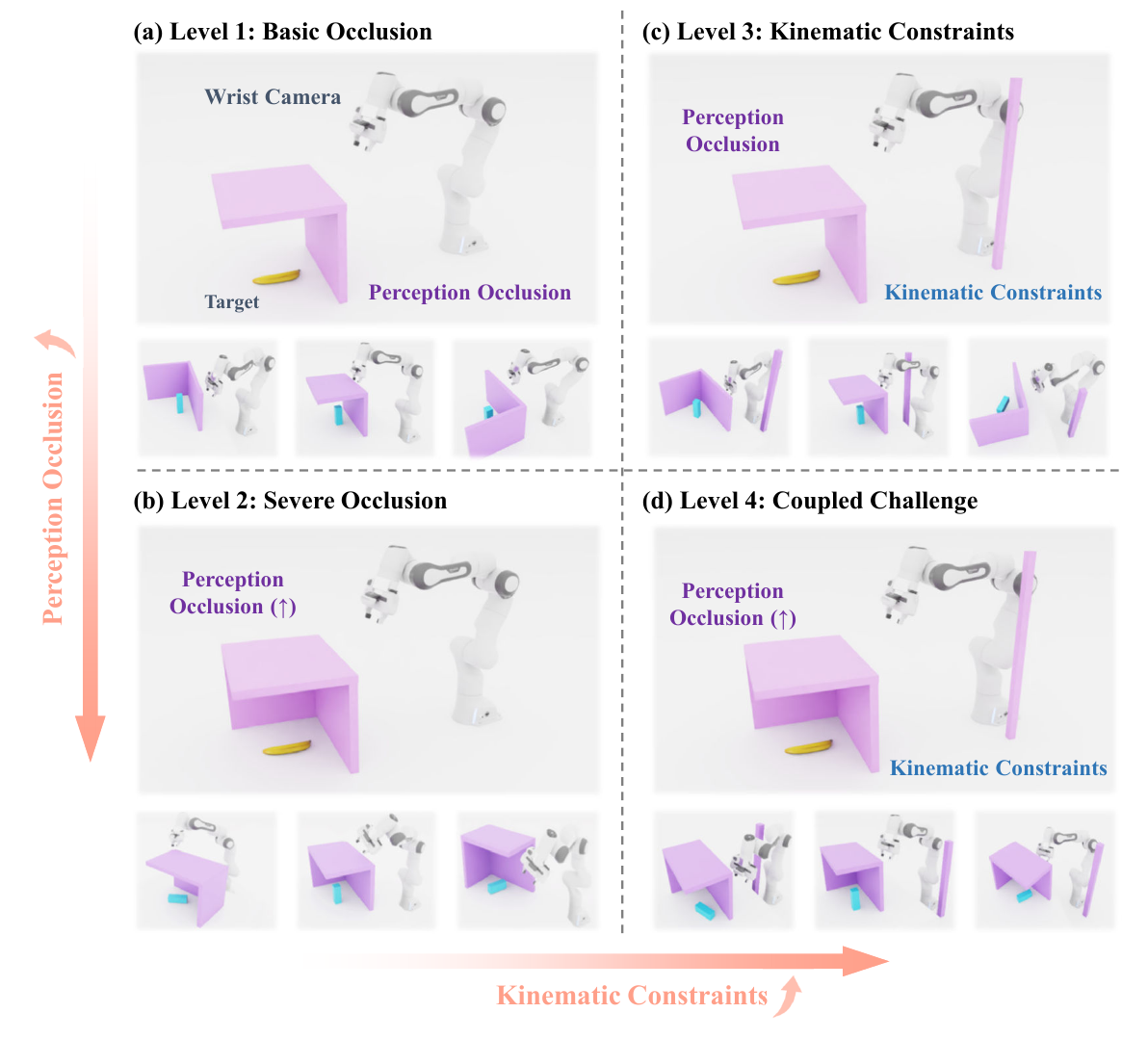}
    \caption{The four challenging scenarios of our proposed benchmark, designed to systematically increase difficulty by combining Perception Occlusion (Levels 1 \& 2) and Kinematic Constraints (Levels 3 \& 4). Level 4 represents the fully coupled challenge.}
    \label{fig:dataset}
    \vspace{-0.3cm}
\end{figure}

\section{Benchmark for Confined-Space Manipulation}
\label{sec:benchmark}

We design a comprehensive benchmark consisting of procedurally generated, confined-space scenes and performance metrics. This benchmark is structured to evaluate an algorithm's ability to handle the two fundamental challenges: perception occlusion and kinematic constraints.

\subsection{Benchmark Scenarios}
We design a benchmark of four scene categories with progressively increasing difficulty. For each category, we procedurally generate 20 unique environments in the Isaac Sim simulator~\cite{NVIDIAIsaacSim}, totaling 80 challenging test cases (Fig.~\ref{fig:dataset}). These categories are constructed to systematically isolate and combine the core challenges, as summarized in Table~\ref{tab:benchmark_levels}.

\textit{Level 1 (Basic Occlusion):} These scenes feature moderate clutter where the target, though occluded by simple obstacles, can be found via monotonic exploration paths.

\textit{Level 2 (Severe Occlusion):} Target accessibility is reduced compared to \textit{Level 1}. The object is hidden, requiring non-monotonic paths (e.g., looking behind an obstacle and then backing out). This category is designed to test an algorithm's ability to find the target through global exploration.

\textit{Level 3 (High Kinematic Constraint):} These scenes feature the same level of occlusion as \textit{Level 1}, but with additional obstacles introduced near the manipulator's base. This setup specifically challenges the planner's awareness of its full-body kinematics. 

\textit{Level 4 (Coupled Challenge):} This level combines the severe occlusion of \textit{Level 2} with the high kinematic constraints of \textit{Level 3}. Success in this level requires an algorithm to plan a global, potentially non-monotonic exploration trajectory to handle severe occlusion, and subsequently conduct manipulation under high kinematic constraints.

\begin{table}[hbtp] 
\centering
\caption{Simulation environment characteristics}
\label{tab:benchmark_levels}
\begin{tabular}{@{}cccc@{}}
\toprule
Level & Perception Occlusion & Kinematic Constraints \\ \midrule
Level 1       & Moderate (+)                  & Low (+)                       \\
Level 2     & High (++)                     & Low (+)                       \\
Level 3      & Moderate (+)                  & High (++)                     \\
Level 4      & High (++)                     & High (++)                     \\ \bottomrule
\end{tabular}

\end{table}

\vspace{-0.4cm}

\subsection{Performance Metrics}

We evaluate performance across three dimensions in solving the coupled challenge in exploration and manipulation.

\begin{itemize}
    \item Exploration Efficiency:
        \begin{itemize}
            \item[\textbullet] Explored Volume over Time: The proportion of the target's bounding box explored versus time. Steeper slopes indicate faster exploration, and higher final values reflect greater coverage.
            \item[\textbullet] Time to Find Target (TFT): The time elapsed until the target is successfully located. This metric reflects exploration efficiency; a lower value is better.
            \item[\textbullet] Path Length to Find Target (PFT): The distance traveled by the end-effector before the target is found; a lower value is better.
        \end{itemize}

    \item Motion Performance:
        \begin{itemize}
            \item[\textbullet] Motion Planning Success Rate (MPSR): The fraction of successful motion planning attempts throughout a task. This metric indicates the kinematic feasibility of the viewpoints; a higher value is better.
            \item[\textbullet] Average Manipulability (AM): The average manipulability index~\cite{yoshikawa1985manipulability} during the exploration process. This metric quantifies the manipulator's distance from singular configurations; a higher value is better.
        \end{itemize}
    
    \item Manipulation Quality: 
        \begin{itemize}
            \item[\textbullet] Grasp Pose Quality (GPQ): A score that reflects the quality of a generated grasp pose, calculated as follows: $\mathcal{S}=s_{\boldsymbol{\mathrm{g}}}\cdot\mathbb{I}(\mathrm{IK}(\boldsymbol{\mathrm{e}}_{\boldsymbol{\mathrm{g}}}) \in \mathcal{Q}_{free})\cdot\cos(\angle(\boldsymbol{z}_{\boldsymbol{\mathrm{g}}}, \boldsymbol{n}))$, where $\mathbb{I}(\cdot)$ is an indicator function that equals 1 if its argument is true, and 0 otherwise. The remaining symbols are defined as in Eq.~\ref{eq:grasp_optimization}. This metric holistically quantifies the grasp's geometric quality and kinematic reachability; a higher value is better.
            \item[\textbullet] Overall Success Rate (SR): The success rate of the entire pipeline, from initial exploration to a successful final grasp. A higher value is better.
        \end{itemize}
\end{itemize}

\section{Experiments}
\label{sec:experiments}

\begin{table*}[t!]
\centering
\caption{Quantitative Performance Comparison Of All Methods In Difficulty Levels.}
\label{tab:full_quantitative_results}
\resizebox{\textwidth}{!}{
\begin{tabular}{@{}ccccccccccccccccc@{}}
\toprule
\multirow{2}{*}{Metric} & \multicolumn{4}{c}{Level1} & \multicolumn{4}{c}{Level2} & \multicolumn{4}{c}{Level3} & \multicolumn{4}{c}{Level4} \\
\cmidrule(lr){2-5} \cmidrule(lr){6-9} \cmidrule(lr){10-13} \cmidrule(lr){14-17}
 & FV & IG & GSE & Proposed &FV & IG & GSE & Proposed & FV & IG & GSE & Proposed & FV & IG & GSE & Proposed \\
\midrule
TFT (s) $\downarrow$ & \textbf{83.5} & 93.9 & 113.7 & 85.7 & 166.9 & 164.3 & 161.5 & \textbf{133.9} & 106.4 & 153.6 & 111.0 & \textbf{95.0} & 173.1 & 173.2 & 158.2 & \textbf{131.4} \\
PFT (m) $\downarrow$ & 4.1 & 4.3 & 4.7 & \textbf{2.9} & 8.4 & 8.2 & 7.1 & \textbf{5.4} & 5.3 & 7.7 & 4.4 & \textbf{3.6} & 8.7 & 8.5 & 7.0 & \textbf{5.5} \\
 MPSR (\%) $\uparrow$ & 66.5 & 39.5 & 59.3 & \textbf{71.3} & 62.5 & 36.4 & 62.3 & \textbf{64.1} & \textbf{61.6} & 34.7 & 60.6 & 60.3 & 59.7 & 41.8 & 59.6 & \textbf{64.9} \\
 AM $\uparrow$ & 0.042 & 0.067 & 0.064 & \textbf{0.070} & 0.033 & 0.062 & 0.065 & \textbf{0.069} & 0.035 & 0.061 & 0.064 & \textbf{0.067} & 0.029 & 0.058 & 0.059 & \textbf{0.066} \\
GPQ $\uparrow$ &  - & 0.578 & 0.670 & \textbf{0.676} & - & 0.544 & \textbf{0.603} & 0.501 & - &0.594  & 0.617 & \textbf{0.634} & - & 0.609 & 0.584 & \textbf{0.613} \\
 SR(\%) $\uparrow$ &- & 40.0 & 47.5 & \textbf{80.0} & - & 27.5 & 42.5 & \textbf{67.0} & - & 30.0 & 47.5 & \textbf{70.0} &  - & 25.0 & 45.0 &\textbf{ 62.5} \\
\bottomrule
\end{tabular}
} 
\vspace{-0.5cm}
\end{table*}

\subsection{Simulation Experiments}
\label{subsec:simulation}

We evaluate our method through simulation experiments in the proposed benchmark environment. These experiments are performed in an Isaac Sim~\cite{NVIDIAIsaacSim} simulation environment, featuring a Franka Panda manipulator equipped with a wrist-mounted depth camera. Motion planning is executed using MoveIt~\cite{chitta2012moveit}, and raw grasp poses are generated by GraspNet~\cite{fang2020graspnet}. All simulations are performed in the Robot Operating System (ROS). We compare our proposed method against the following baselines:

\begin{itemize}
    \item Fixed-Views (FV): A baseline that employs a pre-programmed, open-loop sequence of viewpoints for perception. The resulting trajectory is non-adaptive and does not react to sensory input.
    
    \item Information-Gain Exploration (IG): Based on the method by Isler et al.~\cite{isler2016information}, this approach greedily selects the single viewpoint that maximizes information gain.

    \item Geometric Sampling-based Exploration (GSE): We adapt a sampling-based exploration method from the mobile robotics domain~\cite{zhu2021dsvp, naazare2022online}. This approach utilizes an RRT for spatial expansion but selects the next-best-view based solely on geometric information gain.
\end{itemize}

The performance of all methods is evaluated across the 80 generated scenes, with 10 trials conducted for each scene. A trial concludes when the exploration is reported as complete, the manipulation process finishes, or a predefined time limit is reached. All methods are tested on a computer equipped with an Intel i7-14650HX CPU and a RTX 4060 GPU.

The first two rows of Table~\ref{tab:full_quantitative_results} present the quantitative results for exploration efficiency across all scenarios in our benchmark. Our method consistently achieves higher efficiency in detecting the target, both in terms of time and path length. It is worth noting that since unsuccessful trials are assigned the maximum time limit (i.e., 200s), a lower average time is also indicative of a higher success rate.

\begin{figure}[t]
    \centering
    \includegraphics[width=8.6cm]{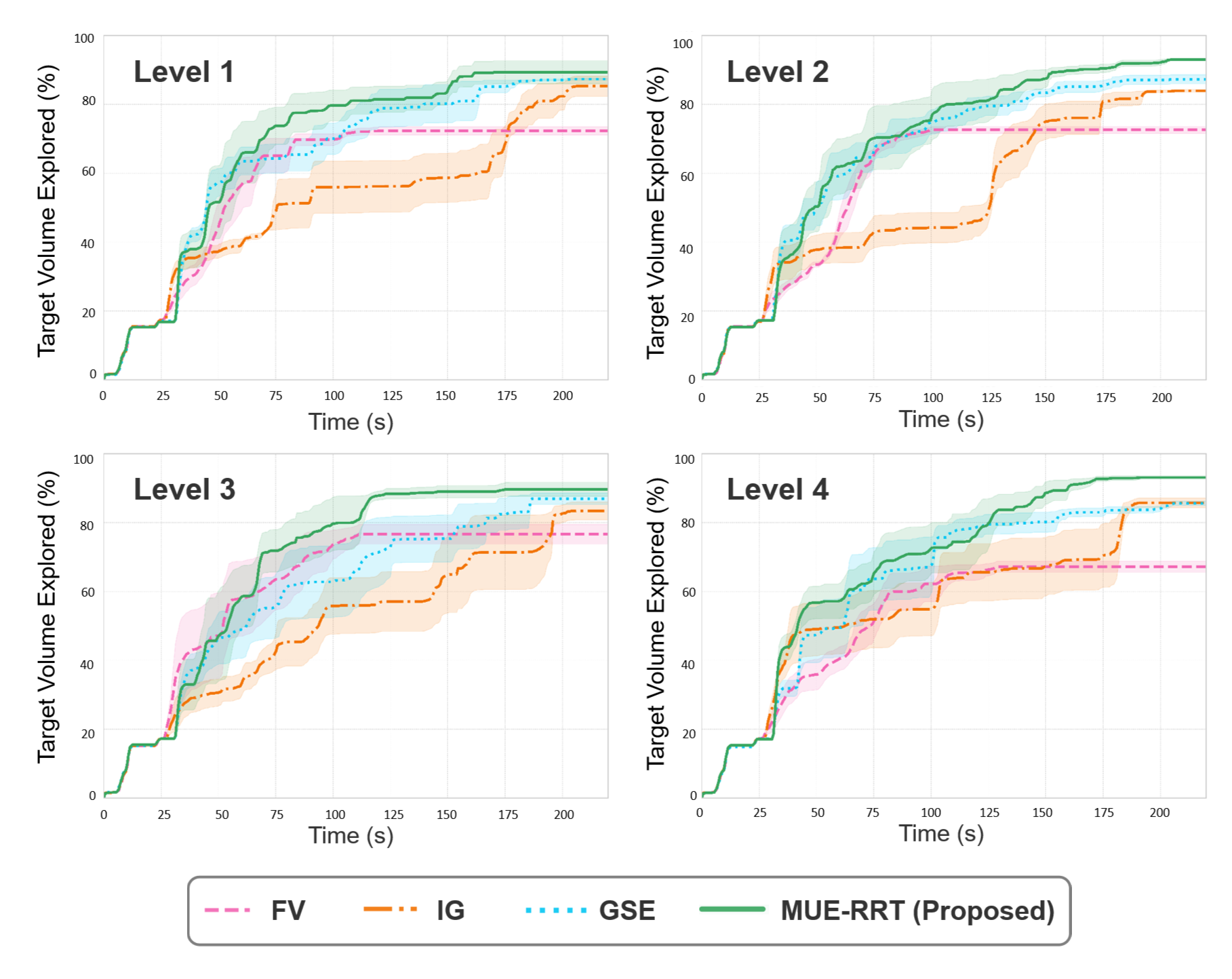}
    \caption{Target volume exploration performance over time across the four benchmark difficulty levels. Solid lines denote the mean value, with shaded areas representing the standard deviation.}

    \label{fig:summary_coverage}
    \vspace{-0.6cm}
\end{figure}

Fig.~\ref{fig:summary_coverage} illustrates the average Explored Volume over Time for each method across the four categories depicted in Fig.~\ref{fig:dataset}. As shown, our approach consistently achieves the highest exploration efficiency. In contrast, FV only attains a fixed and limited coverage, while IG is prone to becoming trapped in local optima and subsequently fails to progress. By integrating RRT with goal-directed guidance, our method enables the most efficient exploration.

The third and fourth rows of Table~\ref{tab:full_quantitative_results} present the quantitative results for motion performance. Compared to the baselines, our method exhibits a higher motion planning success rate during exploration and consistently maintains a higher Average Manipulability. Furthermore, as the environmental kinematic constraints become more complex, the baselines increasingly struggle to find valid motions. This is a direct consequence of their viewpoint selection strategies: the FV is non-adaptive to obstacle configurations, while the Information-Gain based methods greedily pursues perceptual rewards without considering kinematic feasibility.

The last two rows of Table~\ref{tab:full_quantitative_results} present grasp performance. The improvements in exploration and motion directly translate to manipulation quality, with our method achieving an average overall success rate improvement of 39.25\% over IG and 24.25\% over GSE. Since we employed the same constrained pose generation strategy for all methods, the gap in grasp pose quality between the different approaches is relatively small. The primary difference in overall performance stems from the exploration phase.

\subsection{Real-World Experiments}

\begin{figure}[t]
    \centering
    \includegraphics[width=8.6cm]{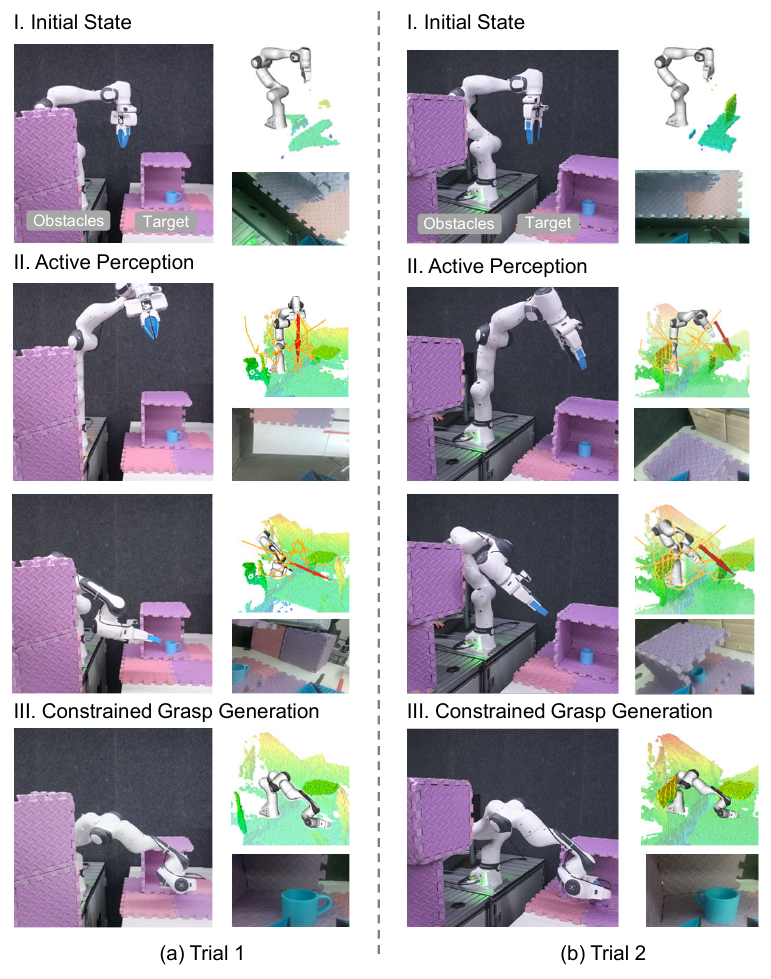}
    \caption{Snapshots of real-world experiments. }
    \label{fig_exp_sen1}
    \vspace{-0.5cm}
\end{figure}

To validate the feasibility and efficiency of our proposed framework, we conduct experiments on a real robotic platform in a confined indoor environment. The platform consists of a 7-DOF Franka Emika FR3 arm equipped with a wrist-mounted RealSense D435i depth camera. We physically recreate a subset of our benchmark scenarios from \textit{Level 4}, which feature both severe perception occlusion and tight kinematic constraints. The robot's task is to locate and retrieve a target object from within a cluttered container, mirroring the objectives of our simulation benchmark. In such environments, planning reactively based on local observations is insufficient and will lead to collisions.

The experimental process is illustrated in Fig.~\ref{fig_exp_sen1}. Initially, the robot has no prior knowledge of the environment (Fig.~\ref{fig_exp_sen1} I). The robot then conducts the active exploration phase, autonomously generating a sequence of viewpoints to actively build a map of the confined space and search for the target (Fig.~\ref{fig_exp_sen1} II). Once the target is detected, the system transitions to the final stage, computing and executing a kinematically feasible, collision-free grasp (Fig.~\ref{fig_exp_sen1} III).

Across 10 trials for each scenario, our method achieves an average Time-to-Find-Target of 159.75s and an overall success rate of 80\%. This high success rate in such challenging, real-world conditions demonstrates the robustness of our integrated perception and planning pipeline. It validates that our exploration strategy can effectively perform active perception to resolve environmental uncertainty, while the subsequent grasp generation module successfully computes and executes constraint-aware grasps. The primary sources of the 20\% failures were twofold: (1) motion planning failures caused by sensor noise, and (2) convergence to local optima in the exploration policy. 

\subsection{Ablation Studies}
\label{sec:ablation}

\subsubsection{Near-Field Awareness for Motion Safety}
To validate the effectiveness of the Near-Field Awareness stage in ensuring motion safety and kinematic feasibility, we compare our full method against a variant in which this stage was disabled. The scenes are chosen from \textit{Level 4}. The results, presented in Table~\ref{tab:ablation_nfa}, show that including the near-field awareness scan significantly increased the motion planning success rate (MPSR) from 54.7\% to 61.2\% and the target detection success rate (DSR) from 68.3\% to 94.1\%, confirming its critical role in ensuring a safe exploration process.

\vspace{-0.2cm}

\begin{table}[hbtp]
\centering
\caption{Ablation study on the Near-Field Awareness}
\label{tab:ablation_nfa}
\begin{tabular}{@{}ccc@{}}
\toprule
Method Variant & MPSR (\%) $\uparrow$ & DSR (\%) $\uparrow$ \\ \midrule
COMPASS (Full Method)      & \textbf{61.2} & \textbf{94.1} \\
w/o Near-Field Awareness   & 54.7         & 68.3  \\ \bottomrule
\end{tabular}
\end{table}

\vspace{-0.2cm}

\subsubsection{Constrained Grasp Pose Optimization for Grasping}
To demonstrate the efficacy of our grasp constraints, we compare our full method against an unconstrained baseline in \textit{Level 4} scenarios. As shown in Table~\ref{tab:ablation_grasp}, omitting these constraints leads to a sharp increase in motion collisions (MC), grasp failures (GF), and object drops (OD), confirming their necessity for safe and successful execution.

\vspace{-0.2cm}

\begin{table}[hbtp]
\centering
\caption{Ablation study on the Constrained Grasp Optimization}
\label{tab:ablation_grasp}
\begin{tabular}{@{}ccccc@{}}
\toprule
Method Variant & Grasp SR (\%) $\uparrow$ & MC $\downarrow$ & GF $\downarrow$ & OD $\downarrow$ \\ \midrule
Full Method     & \textbf{70.0} & \textbf{0/10} & \textbf{1/10} & \textbf{0/10}\\
w/o Grasp Constraints      & 20.0          & 4/10  & 2/10  & 1/10\\ \bottomrule
\end{tabular}
\end{table}

\vspace{-0.5cm}

\section{Conclusion}
\label{sec:conclusion}

In this paper, we have presented COMPASS, a framework for active perception and manipulation in confined spaces. We have proposed the MUE-RRT that ensures safety and motion efficiency. Additionally, we have designed the principled and progressively challenging benchmark for confined-space manipulation tasks. We have conducted extensive simulations and real-world experiments to demonstrate the effectiveness of our method. Compared to exploration methods designed for other robots and only considering information gain, our framework increases manipulation success rate by 24.25\% and reduces the average time to find target by 24.6s in simulations. In the future, we plan to extend this framework to handle dynamic environments via event-depth fusion\cite{luo2024eventtracker} and enable open-vocabulary semantic manipulation\cite{gu2025mr}.


{
    \bibliographystyle{IEEEtran}
    \bibliography{ bib/bibliography}
}

\end{document}